%% file: root.tex
\documentclass[letterpaper, 10pt, conference]{ieeeconf}      

\IEEEoverridecommandlockouts                              

\usepackage{hyperref}       
\usepackage{url}            
\usepackage{booktabs}       
\usepackage{amsfonts}       
\usepackage{nicefrac}       
\usepackage{microtype}      
\usepackage{times}
\usepackage{epsfig}
\usepackage{subcaption, graphicx}
\usepackage{amsmath}
\usepackage{amssymb}
\usepackage{esvect}
\usepackage{algorithm}
\usepackage{algpseudocode}
\usepackage{multirow, multicol}
\usepackage{tabularx}
\usepackage{mathtools}
\usepackage{cite}
\usepackage{color}

\DeclareMathOperator*{\argmax}{arg\,max}

\title{\LARGE \bf
Optimizing Large-Scale Fleet Management on a Road Network using Multi-Agent Deep Reinforcement Learning with Graph Neural Network
}

\author{Juhyeon Kim$^{1, {\dagger}}$ and Kihyun Kim$^{1}$
\thanks{$^{1}$J. Kim and K. Kim are with the Department of Electrical and Computer Engineering, Seoul National University, Seoul 08826, Republic of Korea {\tt\small \{cjdeka3123, hahakhkim\}@snu.ac.kr}}%
\thanks{$^{\dagger}$Corresponding author} 
}

\begin{document}

\maketitle
\thispagestyle{empty}
\pagestyle{empty}

\begin{abstract}
We propose a novel approach to optimize fleet management by combining multi-agent reinforcement learning with graph neural network. 
To provide ride-hailing service, one needs to optimize dynamic resources and demands over spatial domain.
While the spatial structure was previously approximated with a regular grid, our approach represents the road network with a graph, which better reflects the underlying geometric structure.
Dynamic resource allocation is formulated as multi-agent reinforcement learning, whose action-value function (Q function) is approximated with graph neural networks.
We use stochastic policy update rule over the graph with deep Q-networks (DQN), and achieve superior results over the greedy policy update.
We design a realistic simulator that emulates the empirical taxi call data, and confirm the effectiveness of the proposed model under various conditions. 
\end{abstract}

\input{01_introduction.tex}
\input{02_related_works.tex}

\input{03_problem_statement.tex}
\input{04_MARLwithGNN.tex}
\input{05_simulator_design.tex}
\input{06_experiments.tex}
\input{07_conclusion.tex}
\input{08_appendix.tex}
\addtolength{\textheight}{0cm}   




\section*{ACKNOWLEDGMENT}
We appreciate Seungil You at \textit{Kakao Mobility} for constructive advice.
We would also like to thank Young Min Kim at Seoul National University for revising the manuscript.


\small
\bibliographystyle{IEEEtran}
\typeout{}
\bibliography{reference}
\end{document}

%% file: 01_introduction.tex
\section{Introduction}
\label{sec:intro}
One of the main challenges in ride-hailing services is supply and demand balancing—making vehicles available at the right place and time for customers.
To achieve this goal, establishing a smart and timely reallocation strategy for dynamic transportation resources is crucial, which is known as a \textit{fleet management problem}.
Without such efforts, customers will suffer from the lack of drivers, while some drivers will remain idle in near areas looking for orders.
Though there exist rich empirical demand and supply data, designing an optimal vehicle management system still remains a challenging task.
One of the main obstacles is the dynamic nature of the fleet management problem that the current relocation policy affects the future demand-supply distribution.

To resolve such difficulty, reinforcement learning (RL) has been widely adopted in previous works \cite{godfrey2002adaptive, wei2017look}.
In RL frameworks, agents iteratively learn a strategy to maximize the reward by interacting with an environment.
To establish a strategy that has practical implications, an accurate but manageable problem setting and RL framework design is critical.
To this end, recent studies \cite{lin2018efficient, li2019efficient, zhou2019multi, ke2019optimizing} suggested multi-agent deep reinforcement learning (MADRL) to effectively approximate highly complicated fleet management scenario.
Though these works have brought an important insight, they still have limitations to be applied in real systems since they oversimplified the spatial structure by assuming a grid-shaped environment.
If a graph is used to model the environment, it would be possible to allocate drivers more precisely at the road level.
Yet using a graph with previous methods is difficult since a graph requires a relatively complex structure whose components are connected in an irregular manner.

This paper proposes a novel approach to a fleet management problem using MADRL with graph neural network (GNN).
Our study is based on the aforementioned papers \cite{lin2018efficient, li2019efficient, zhou2019multi, ke2019optimizing}, while we modify their settings to fit with our graph model of a road network.
Specifically, while drivers in the grid-based approaches should only move to one of the neighboring grid cells, drivers in our model can move to any connected roads.
Moreover, the exact location of each driver can be pinpointed in our model using the relative position on its current road, which was impossible in the previous works.
As a result, our model can handle each order and driver more precisely at the road level.
To find an effective relocation policy, we suggest a novel method using GNN to precisely estimate an action-value function (i.e. $Q$ function). 
Inspired by the fact that the $Q$ function of each road is highly dependent on its connected roads, we transform an entire road network to the appropriate GNN model and train this model using DQN.
Finally, we suggest a  stochastic policy update from $Q$ function to efficiently handle a large number of drivers.
Our observation shows that this improves the overall performance compared to conventional $\epsilon$-greedy policy, by reasonably distributing drivers according to the $Q$ function.

%% file: 02_related_works.tex
\section{Related Work}
\label{sec:related_works}
\subsection{Large-scale fleet management}
A fleet management problem has been extensively studied in a rule-based manner.
For example, there exist works using greedy matching \cite{lee2004taxi}, centralized combinatorial optimization \cite{zhang2017taxi}, collaborative dispatch with decentralized setting \cite{seow2009collaborative}, or adaptive multi-agent scheduling system \cite{alshamsi2009multiagent}.
To exclude heuristics in the rule-based model, \cite{xu2018large} and \cite{wang2018deep} adopted an RL-based approach, but they assumed a single-agent setting that cannot model the complex interactions between large numbers of drivers and orders.

The more recent trend is the adoption of MADRL.
\cite{lin2018efficient} proposed a contextual MADRL that considers geographic and collaborative contexts to prevent invalid actions. 
\cite{li2019efficient} adopted mean-field approximation to model agents' interaction and \cite{zhou2019multi} proposed a decentralized system based on MADRL with Kullback–Leibler divergence optimization.
Besides, \cite{ke2019optimizing} used delayed order matching with a two-stage framework utilizing both MADRL and combinatorial method.
However, these studies assumed a grid system which is highly different from real roads, which is represented as a graph.
Though there exist several studies dealt with graph combinatorial optimization with RL \cite{khalil2017learning, bello2016neural}, a large-scale problem using MADRL has been rarely studied.

\subsection{Multi-agent reinforcement learning}
In multi-agent reinforcement learning (MARL) frameworks, a large number of agents interact with the same environment and receive rewards corresponding to their state and action.
\cite{tan1993multi} is one of the first papers that dealt with a multi-agent scheme, proposing a cooperative $Q$-learning that agents share their policy and experience.
\cite{foerster2018counterfactual} used counterfactual multi-agent policy gradient model which comprised of centralized critic and decentralized actor. 
\cite{lowe2017multi} also used similar approach, but added extra information in training phase.
However, these works are limited to a handful number of agents due to communication costs.
\cite{zheng2018magent} alleviated this complexity by a homogeneity assumption and introduced a highly scalable multi-agent simulation platform called MAgent. 
\cite{yang2018mean} further developed MAgent with \textit{mean-field} theory that approximates interactions between agents, and showed it converges to the Nash equilibrium.

%% file: 03_problem_statement.tex
\section{Problem Statement}
\label{sec:problem_statment}
In this section, we specify a problem setting of the fleet management problem.
Overall problem settings and notations are similar to those of previous works \cite{lin2018efficient, li2019efficient, zhou2019multi, ke2019optimizing}.
The major difference is a spatial setting—we use a graph rather than a grid.
Also we use the total order response rate, number of served orders divided by the number of total orders, as an objective to be maximized, instead of using the gross merchandise volume (GMV) that has been used in previous works.
The advantage of using the order response rate over GMV is that it can reflect the customers' satisfaction as well as the revenue.
 
\subsection{Graph representation of a road network}
\begin{figure}[t]
\begin{center}
\includegraphics[width=0.6\linewidth]{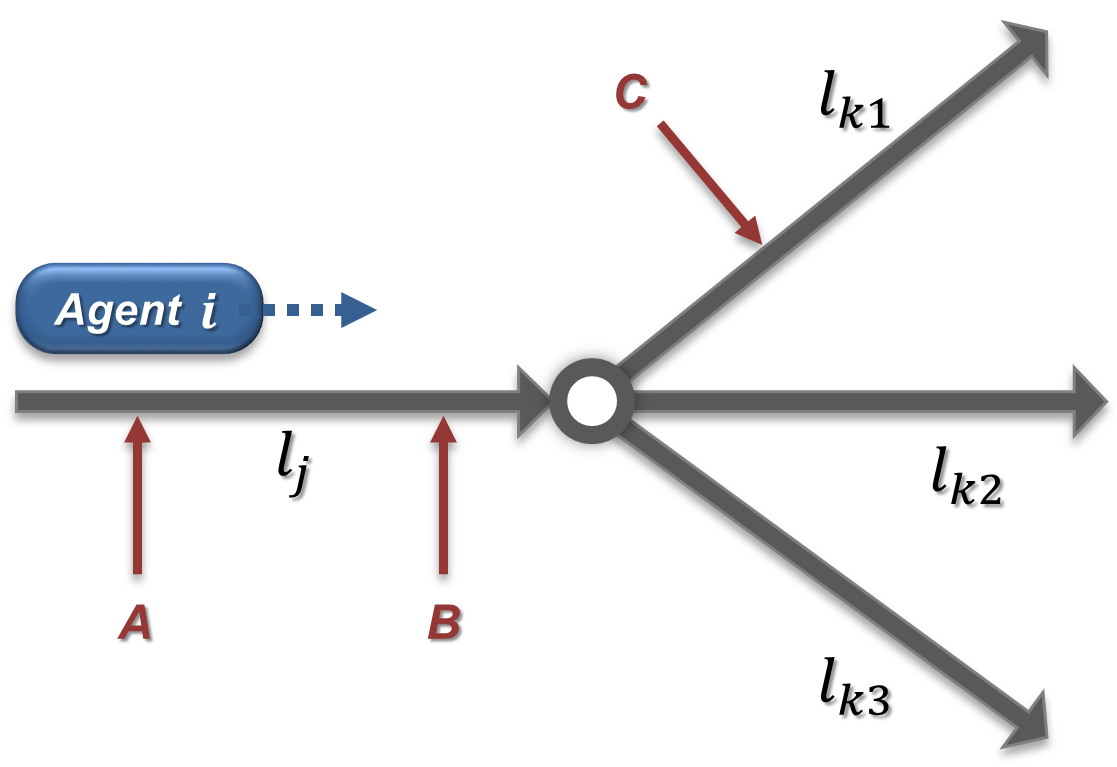}
\end{center}
   \caption{Illustration of a road network. The agent $i$ is currently located at the position $A$ on the road $l_j$. At the next time step, it moves forward. Depending on the speed of the road $l_j$, it can either stay on the same road (position $B$) or move on to one of the neighboring roads $l_k$s (position $C$).}
\label{fig:agent_on_road_network}
\end{figure}

 A road network can be modeled as a directed graph $G_R :=(V, E)$, where $V$ denotes the set of intersections and $E:=\{l_j \mid j=1, 2, \ldots, N_{\text{road}}\}$ denotes the set of roads.
 Drivers are distributed throughout the road network (Fig. \ref{fig:agent_on_road_network}). 
 Note that they are not on the nodes, but on the edges (roads). 
 The location of any driver can be represented by the road index and the relative position on the road.
 
 At each time step, all drivers move forward by the speed of their current road.
 Staying at the current position is not allowed in our setting.
 If a driver cannot reach the end of the current road, thereby it cannot transit to a different road within the current time step, we consider it as a \emph{non-controllable} driver.
 Conversely, if a driver can reach the end of the current road, it is regarded as a \emph{controllable} driver who can move on to the next roads.
 Each controllable driver should be relocated to one of its connected roads.
 We randomly set the relative position of relocated agent on the new road, according to the uniform distribution.
 It reflects the time spent at the previous intersection and also gives randomness.
 We also limit the number of transitions to 1 per time step, instead shorten the unit time step to make the movable distance per time step short enough. 
 Orders (calls) are generated from each road at each time step and are randomly assigned to idle drivers at the same road, those who do not currently serve any order.
 Under this setting, our ultimate goal is to find an optimal relocation strategy for controllable drivers that maximizes the order response rate.

\subsection{Markov game modeling}
Now we define a Markov game $G:=(N, \mathcal{S, A, P, R}, \gamma)$.
$(N, \mathcal{S, A, P, R}, \gamma)$ refers to the total agent number, state space, joint action space, transition probability, reward function, and discount factor, respectively.
\begin{itemize}
    \item Agent : We define an agent as a driver who is at the idle state, which can be either controllable or non-controllable.
    Controllable agents at the same road and the time are assumed to be homogeneous, implying that they share the same policy. Orders are randomly assigned to both types of agents regardless of their relative position on the road.
    Let $N_{t}$ denote the number total agents at time $t$.
    We denote the road that $i$th agent is located on at time $t$ as $l_t^i$, where $i=1, 2, \ldots, N_t$. 
    We also define $A_t^c$ as the set of all controllable agents at time $t$.
    \item State $s_{t} \in \mathcal{S}$: We use the global state $s_t$ at each time $t$. We observe three values at each road $l_j$ at time $t$: the number of agents (idle drivers) $N_{j, t}$, the number of calls $N^{\text{call}}_{j, t}$, and the speed $speed_{j, t}$.
    We concatenate this information for all roads to obtain the global state
    \begin{equation}
        s_{t}:= [(N_{j, t},N^{\text{call}}_{j, t}, speed_{j, t})]_{j=1}^{{N_{\text{road}}}} \in \mathbb{N}_{0}^{{N_{\text{road}}}\times 2} \times \mathbb{R}^{N_{\text{road}}}.
    \end{equation}
    \item Action $a_{t} \in \mathcal{A}$: 
    We define the action $a_t^i$ as the decision of the next road for the agent $i$ at time $t$. Note that each road has a different set of possible actions.
    We will indicate the action moving from $l_j$ to $l_k$ as $l_j\rightarrow l_k$.
    For non-controllable agents, their action is limited to stay on the current road (e.g. $l_j \rightarrow l_j$).
    We aggregate all agents' action to define a joint action $a_{t}:=[a_{t}^{i}]_{i=1}^{N_t}$.
    \item Reward $\mathcal{R}_t \in \mathcal{R}$: $\mathcal{R}_t := [\mathcal{R}^i_t]_{i=1}^{N_t}$, where $\mathcal{R}^i_t$ is defined as the $i$th agent's reward at time $t$.
    After the action $a_{t}^{i}$, $R^i_{t}$ is simply set to $1$ if an order is assigned or $0$ if not. Note that non-controllable drivers can also serve calls and receive the reward.
    \item State transition probability $\mathcal{P}$:
    The action is deterministic, but since the number of drivers varies in the real world due to commuting, we add or remove idle drivers on random positions so that total number of idle drivers at time $t$ fit with the historical data.
\end{itemize}

The $i$th agent's discounted return at time $t$ is given by $G^{i}_{t}:=\sum_{k=0}^{\infty}\gamma^{k}R^i_{t+k}$, where $\gamma \in (0, 1)$ is a discount factor.
To consider the road-level policy, we define $\pi (l_j \rightarrow l_k | s_t)$, which denotes the probability of the action $l_j \rightarrow l_k$ at the road $l_j$.
By definition, $\sum_{l_k \in S(l_j)} \pi (l_j \rightarrow l_k | s_t) = 1$ for $\forall s_t, \forall j$, where  $S(l_j)$ denotes the set of all successor roads of $l_j$.

\subsection{Bellman expectation equation}
We define the state-value function and the action-value function ($Q$ function) as follows:
\begin{equation}
\begin{split}
    &V^{\pi}(s_t, l_j) := \mathbb{E}^{\pi}[G^{i}_{t}\mid \text{agent $i$ is on $l_j$ at time $t$ before} \\
    &\hspace{8em} \text{movement, under the state $s_t$}],\\
    & Q^{\pi}(s_t, l_j \rightarrow l_k) := \mathbb{E}^{\pi}[G^{i}_{t} \mid \text{agent $i$ moves $l_j\rightarrow l_k$}\\
    &\hspace{10em} \text{at time $t$, under the state $s_t$}].
\end{split}
\end{equation}
Here, $\mathbb{E}^\pi$ denotes the expectation value when all agents move according to the policy $\pi$.
Note that our state-value function $V^\pi(s_t,l_j)$ is marginalized over the controllability of all agents on the road $l_j$.
In other words, $V^\pi(s_t,l_j)$ is an expected value for both controllable and non-controllable agents.

To simplify our formulation, we need two assumptions.
First, we assume that the expected reward does not depend on the agent's relative position on the road.
Thus, the expected reward of any agent on the road $l_j$ is identical as $V^\pi(s_t,l_j)$ regardless of its relative position on $l_j$.
Second, we assume that $Q^{\pi}(s_t, l_j \rightarrow l_k)$ does not depend on the departed road $l_j$ (as in \cite{lin2018efficient}), which allows us the approximation $
Q^{\pi}(s_t, l_j \rightarrow l_k) \approx Q^{\pi}(s_t, l_k), \; \forall j=1,2, \ldots, N_{\text{road}}$.
Though we cannot take account of the cost of moving from $l_j$ to $l_k$ under this assumption, we can greatly reduce the complexity of problem.
Remark that approximated $Q^{\pi}(s_t, l_k)$ has the same input-output structure to the state-value function $V^{\pi}(s_t, l_j)$.


The state-value function can be expressed as follows by dividing into non-controllable case and controllable case:
\begin{equation}\label{eq:bellman1}
\begin{split}
    & V^{\pi} (s_t, l_j) \approx (1-p^c_{j,t}) Q^{\pi} (s_{t}, l_j)\\
    &\hspace{4em}+  p^c_{j,t} \sum_{l_k \in S(l_j)} \pi (l_j \rightarrow l_k| s_t)Q^{\pi}(s_{t}, l_k),
\end{split}
\end{equation}
where $p^c_{j,t}$ denotes the probability that an agent on the road $l_j$ at the time step $t$ is controllable.
The Bellman expectation equation is then given by
\begin{equation}\label{eq:bellman2}
\begin{split}
    &Q^{\pi} (s_t, l_k) = \mathbb{E}[R^i_{t} \mid \text{agent $i$ is on $l_k$ at time $t$ after}  \\
    &\hspace{4em}\text{movement, under the state $s_t$}] + \gamma V^{\pi} (s_{t+1}, l_k).
\end{split}
\end{equation}
Equations \eqref{eq:bellman1} and \eqref{eq:bellman2} will be used to estimate $Q$ function.

%% file: 04_MARLwithGNN.tex
\section{Multi-agent Reinforcement Learning with Graph Neural Network}
\label{sec:MARL_with_GNN}

\begin{figure*}[t!]
\centering
\begin{subfigure}[b]{0.30\textwidth}
  \centering
  \includegraphics[width=\linewidth]{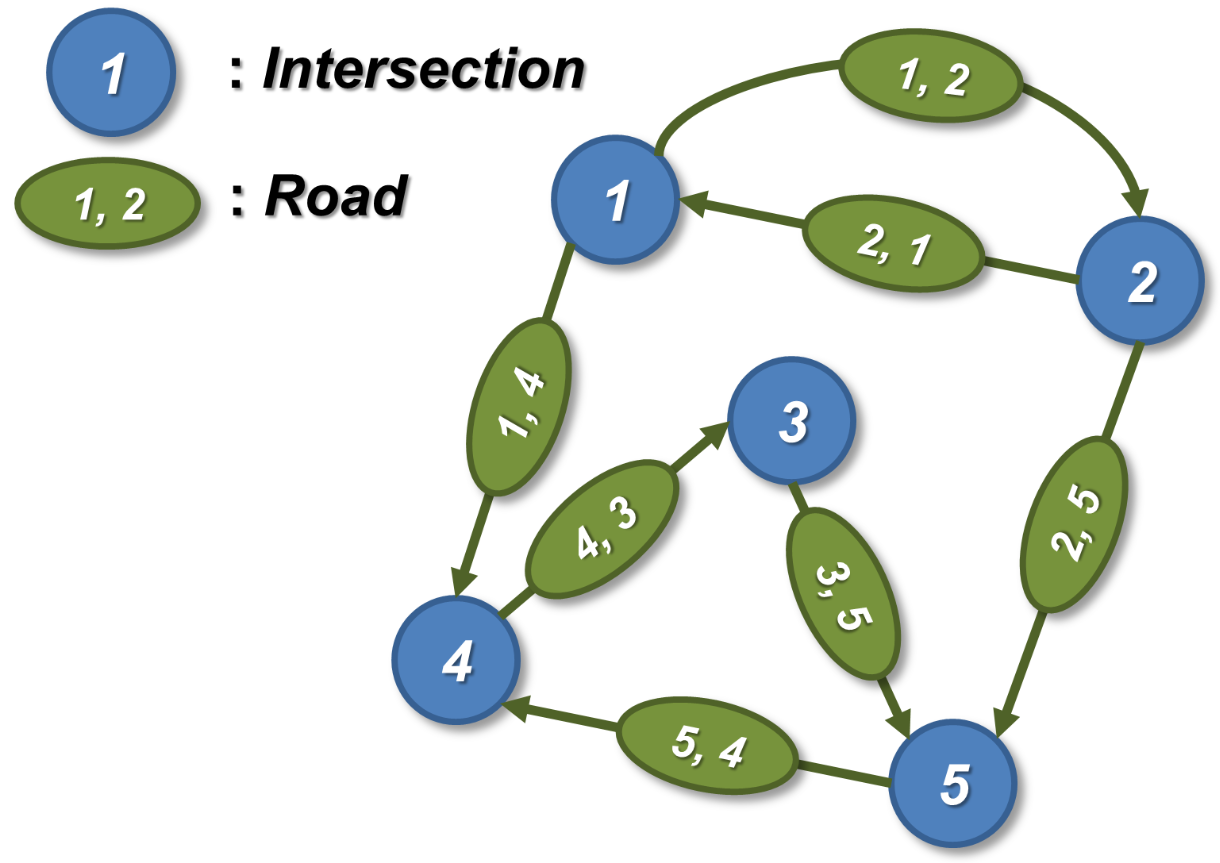}
  \caption{}
\end{subfigure}
\hspace{0.5em}
\begin{subfigure}[b]{0.21\textwidth}
  \centering
  \includegraphics[width=\linewidth]{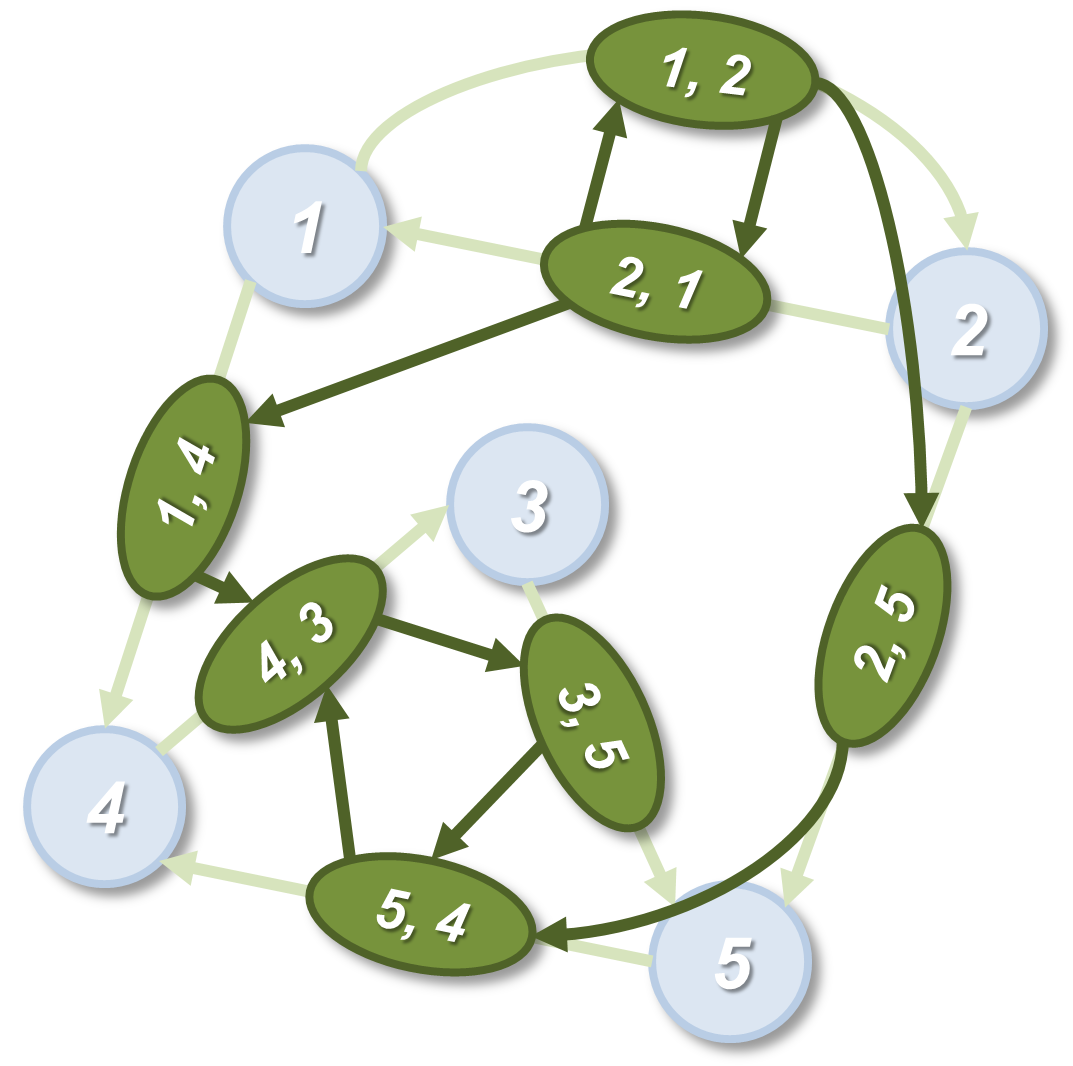}
  \caption{}
\end{subfigure}
\hspace{0.5em}%
\begin{subfigure}[b]{0.18\textwidth}
  \centering
  \includegraphics[width=\linewidth]{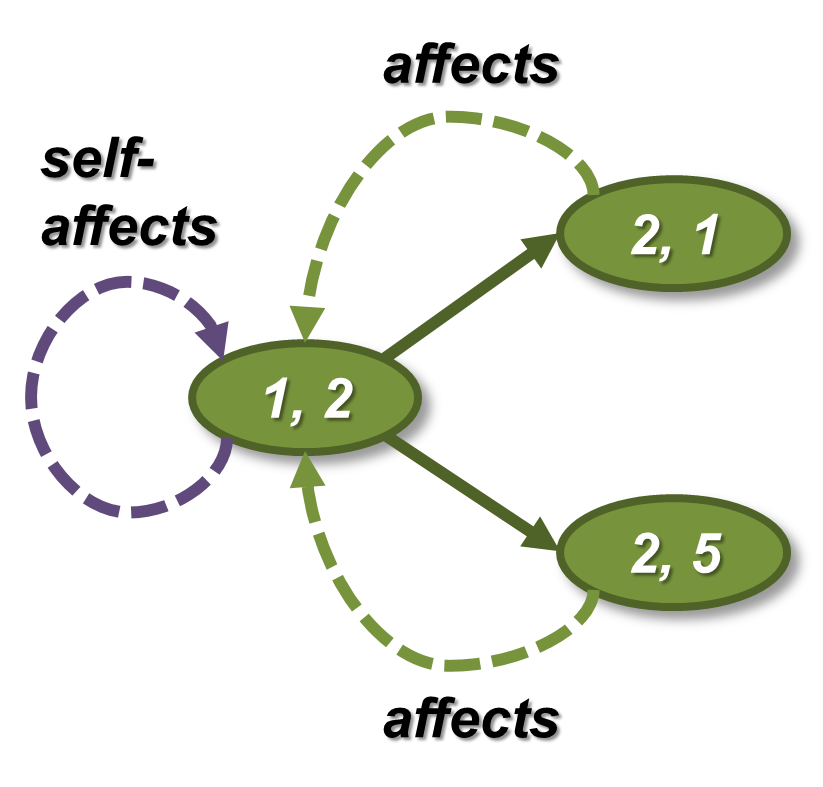}
  \caption{}
\end{subfigure}
\hspace{0.5em}%
\begin{subfigure}[b]{0.19\textwidth}
  \centering
  \includegraphics[width=\linewidth]{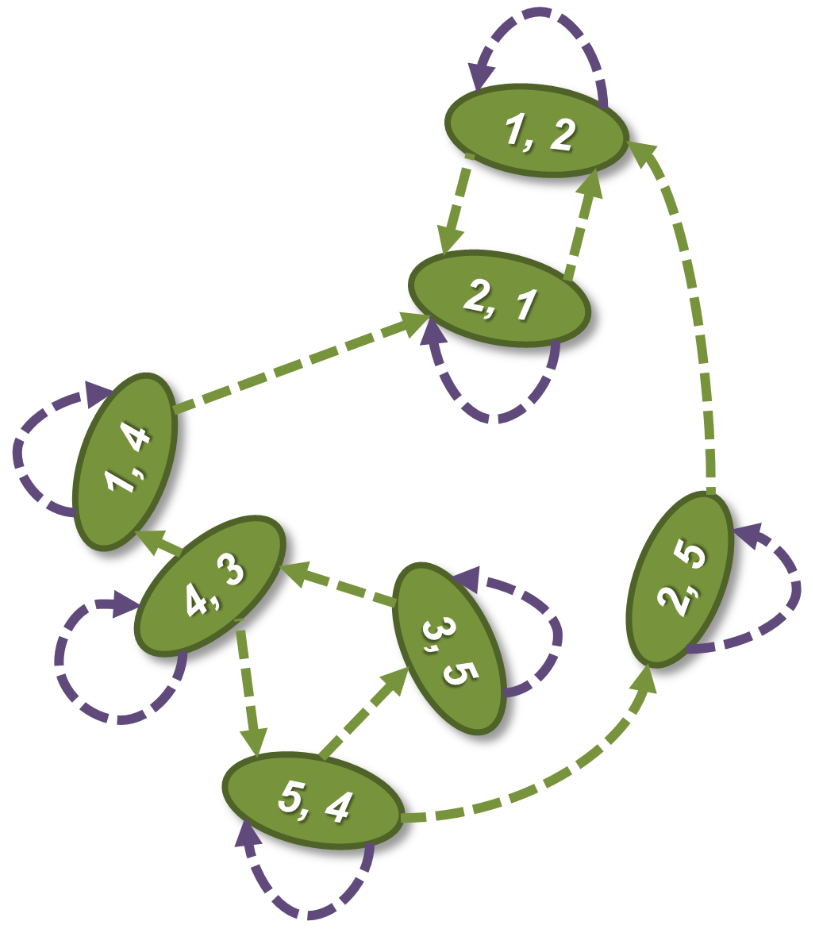}
  \caption{}
\end{subfigure}
\caption{Converting road network $G_R$ to a proper graph form to apply graph neural networks: (a) original road network, (b) line graph conversion, (c) adding self-dependency and reversing direction, (d) the final graph after conversion.}
\label{fig:graph_conversion}
\end{figure*}

\subsection{Large-scale multi-agent reinforcement learning}
We use $Q$-learning to estimate the $Q$ function in our MARL problem as in \cite{lin2018efficient, zhou2019multi, li2019cooperative}.
In conventional RL, an optimal policy ${\pi}^{\ast}$ satisfies the following greediness property, where $Q_{\ast}$ is an optimal action-value function:
\begin{equation}
        {\pi}^{\ast}(a|s)= 
\begin{dcases}
    1  & \text{if    }a = \argmax_{a \in \mathcal{A} }Q_{\ast}(s, a)\\
    0  & \text{otherwise}.
\end{dcases}
\end{equation}
However, in a large-scale multi-agent problem, this greediness property does not hold unless we control agents one by one.
For example, in fleet management problem, it is intuitively clear that sending all agents to the next road whose $Q$ value is the highest, will not be optimal in most cases.

As a result, an optimal policy for our problem setting would be stochastic, not all-or-nothing.
\cite{lin2018efficient} dealt with this by simply keeping a small portion of randomness with the $\epsilon$-greedy policy ($\epsilon=0.1$) during evaluation.
However, this approach has a limitation that only the maximum $Q$ value is considered in the stochastic policy.
Instead, we propose a method that does not rely on the optimal policy greediness or the Bellman optimality equation.
We replace $Q$ function update in the standard $Q$-learning
to the following update rule, which is known as \emph{expected-SARSA} \cite{sutton1998introduction}:
\begin{equation}\label{eq:modified_q_value_update}
\begin{split}
   Q(s,a) \leftarrow &(1-\alpha) Q(s,a)\\
   &+ \alpha\left(R(s, a, s') + \gamma \mathbb{E}[Q(s', a')]\right),
\end{split}
\end{equation}
where $\mathbb{E}[Q(s', a')] = \sum_{a'\in \mathcal{A}}{\pi}(a'|s')Q(s', a')$.
A policy update rule is also modified in our method.
Unfortunately, it is difficult to find an optimal stochastic policy update method in a complex model such as a fleet management problem.
One plausible setting is to increase the probability of the action which has a large $Q$ value.
We can simply achieve this by setting ${\pi}(a|s) = F_{\pi}(Q(s, a))$, 
where $F_{\pi}$ is an increasing function of $Q(s,a)$ satisfying $\sum_{a \in \mathcal{A}} F_{\pi}(Q(s,a)) = 1$.
There are countless options for $F_{\pi}$, but here, we simply set it to normalized $\beta$-squared function or $\beta$-exponential function as follows:
\begin{equation}
    {\pi}(a|s) = \frac{{Q(s, a)}^\beta}{\sum_{a \in \mathcal{A}}{Q(s, a)}^\beta} \text{ or } \frac{\exp(\beta {Q(s, a)})}{\sum_{a\in \mathcal{A}}\exp(\beta {Q(s, a)})}.
\label{eq:modified_policy_update}
\end{equation}
Note that above converges to a greedy policy as $\beta \rightarrow \infty$.

Overall value/policy update is similar to that of maximum entropy based approaches or soft $Q$-learning\cite{haarnoja2017reinforcement}.
The major difference is the presence of an entropy term.
Soft $Q$-learning encouraged the stochastic behavior by introducing an additional entropy term.
However in our problem setting, since the greedy policy leads to the immediate reduction of reward, there is no reason to add an entropy term for exploration.
\emph{Mean-field} in \cite{li2019efficient} is also similar to our approach, but they used the stochastic policy in the perspective of exploration and set greedy at the end of the learning, which is different from our usage.
There is no guarantee that our update method converges to an optimal policy, but heuristically we can expect a better result than using a greedy policy.

Now, we reformulate $Q$ function update rule \eqref{eq:modified_q_value_update} with our problem settings and notations.
Consider updating $i$th agent's experience whose state transition sample is given by $(s_t, a_t^i=l_t^i \rightarrow l_{t+1}^i, R_t^i, s_{t+1})$.
From \eqref{eq:bellman1} and \eqref{eq:bellman2}, our update rule can be expressed as
\begin{equation}
\begin{split}
    Q^\pi(s_t, l_{t+1}^i) \leftarrow & (1-\alpha) Q^\pi(s_t, l_{t+1}^i)\\
    &+ \alpha \left[ R^i_t + \gamma \hat{Q}^\pi(s_{t+1},l_{t+1}^i, i) \right],
\end{split}
\label{eq:our_q_value_update}
\end{equation}
where $\hat{Q}^\pi(s_{t}, l_j, i)$ is
\begin{equation}
\begin{split}
\begin{dcases}
    \sum_{l_k \in S(l_j)}{\pi} (l_j \rightarrow l_k | s_t) Q^\pi(s_t, l_k)  & \text{if } i \in A^c_t\\
    Q^\pi(s_t, l_j)  & \text{if } i \not\in A^c_t.
\end{dcases}
\end{split}
\label{eq:our_q_value_update_T}
\end{equation}
In \eqref{eq:our_q_value_update_T}, the case $i \in A^c_t$ denotes the expected future value of $Q$ function when the agent $i$ is controllable, and the case $i \not\in A^c_t$ denotes the value when $i$ is non-controllable. 
Remark that $\mathbb{E}[\hat{Q}^\pi(s_{t}, l_j, i)] = V^\pi(s_t, l_j)$ since $V^\pi(s_t, l_j)$ is marginalized over the controllability by definition.
Our stochastic policy can be rewritten as follows using \eqref{eq:modified_policy_update}:
\begin{equation}
    {\pi} (l_j \rightarrow l_k | s_t) = \frac{{Q^\pi(s_t, l_k)}^\beta}{\sum_{l_k \in S(l_j)}{Q^\pi(s_t, l_k)}^\beta},
\label{eq:our_policy_update}
\end{equation}
which is similar for the exponential case.

We adopt the methodology of DQN \cite{mnih2013playing} but modify it slightly to suit our problem setting.
The mean squared error (MSE) for $i$th agent at time $t$ is given by
\begin{equation}
    \left[Q^\pi(s_t, l_{t+1}^i;\theta) - \left\{R^i_t + \gamma \hat{Q}^{\pi'}(s_{t+1}, l_{t+1}^i, i; {\theta}')\right\}\right]^2,
    \label{equ:mse_dqn}
\end{equation}
where $\hat{Q}^{\pi'}(s_{t+1}, l_{t+1}^i, i; {\theta}')$ is defined as \eqref{eq:our_q_value_update_T} using $Q', {\pi}'$ calculated from the target neural network with parameter ${\theta}'$.
Our goal is to minimize the above MSE loss of all agents.
For simplicity, we set the driver's state to be terminated if it newly serves an order.
We do not use a replay memory, because the correlation of experiences from multiple agents is already diluted.

\subsection{Graph neural network as a function-approximator}
To approximate the $Q$ function, we use a graph neural network (GNN).
GNN takes a directed graph $G=(V,E)$ and a $d$-dimensional graph signal $X\in\mathbb{R}^{|V|\times d}$ as an input.
We use two basic models, GCN \cite{kipf2016semi} and GAT \cite{velivckovic2017graph}. 
For both GCN and GAT, a graph convolution operation of $i$th node at $l$th layer can be expressed as $h_{i}^{(l+1)} = \sigma{(\sum_{j\in \mathcal{P}(i)}c W^{(l)}h_{j}^{(l)})}$,
where $h_i^{(l)}$ is a $l$th layer node embedding, $\mathcal{P}(i)$ is a set of node $i$'s predecessors, $\sigma$ is an activation function, $W^{(l)}$ is a learnable weight matrix for node-wise feature transformation and $c$ is a normalizing constant which is constant (GCN) or learned with attention (GAT).

 We aim to design GNN that takes $s_t\in\mathbb{R}^{{N_{\text{road}}}\times 3}$ as an input and generates $Q(s_t)\in\mathbb{R}^{{N_{\text{road}}}}$ as an output.
 To this end, we transform the road network $G_R$ as illustrated in Fig. \ref{fig:graph_conversion}.
 First, we convert $G_R$ to its edge-vertex dual graph, or a line graph, $L(G_R)$ (Fig. \ref{fig:graph_conversion}(b)), because each road must be considered as a node.
 Now consider the two components of the update target in $Q$ function: reward term $R$ and discounted future return $\hat{Q}$.
 The reward term is affected by the current road, while the discounted future return is affected by its successors.
 Note that in both GCN and GAT, the message is passed to successors, meaning that $h_{i}^{(l)}$ is computed by its predecessors.
 Thus, in order to make successor roads affect predecessor roads, we need to reverse the direction of all edges (Fig. \ref{fig:graph_conversion}(c)).
 We also need to add a self-loop for each node to include the self-dependency required for computing the reward term.
 Fig. \ref{fig:graph_conversion}(d) shows the final shape of the graph after conversion.

%% file: 05_simulator_design.tex
\section{Simulator Design}
\label{sec:simulator design}
 
 Here, we introduce how we design the simulator that models real ride-hailing services \footnote{ The code is available at \url{https://github.com/juhyeonkim95/TaxiSimulatorOnGraph}}.
 To run the simulator, we need call data within a period, the number of total drivers over time, and initial idle driver distribution.
 At the initialization step, we deploy idle drivers at each road $l_j$ corresponding to the initial idle driver distribution and then assign calls.
 At each time step $t$, we go through the following cycle taking the policy $\pi_t$ as input and returning new observation $s_{t+1}$. (i) Move every driver forward by the current speed of each road. If one can move further than the end of its current road, add it to the controllable driver list. (ii) Relocate all controllable drivers to the next road by applying policy ${\pi}_t$. Also update non-controllable drivers' position. (iii) Assign orders to the drivers on the same road. (iv) Increase time step by 1 and check whether each driver has finished current job or not. (v) Generate orders at each road $l_j$ using the data $Call(j, t)$. Each call data is given by a tuple of (start road, end road, start time, duration, cost). (vi)  Generate or remove drivers to fit the given number of total drivers $N_{t}^{\text{total}}$, to consider drivers being offline or online. In addition, remove expired orders and set the new speed for each road. (vii) Observe the next state $s_{t+1}=[(N_{j, t+1}, N^{\text{call}}_{j, t+1}, speed_{j,t+1}) ]^{N_{\text{road}}}_{j=1}$.

%% file: 06_experiments.tex
\section{Experiments}
\label{sec:experiments}
\subsection{Experimental setting}\label{sec:setting}
\newcommand\city{Seoul}
In this section, the performance of the proposed models is demonstrated and compared under various conditions.
We trained and evaluated our algorithm using aggregated and anonymized data in simplified Seoul road network, provided by \emph{Kakao Mobility}.
The speed data of each road is gathered from the public website.
We split one day into 1440 steps (1 minute for each step), trained and tested for 5 epochs (days).
We used 8 layers for both GCN and GAT. For GAT, we used 8 attention heads. ReLU is used as an activation function except the last layer that used sigmoid to guarantee $Q$ value between 0 and 1.
We balanced exploration/exploitation by setting policy to $(1-\epsilon)\pi + \epsilon {\pi}_{r}$, where ${\pi}$ denotes our computed policy, and ${\pi}_r$ denotes a uniform random policy.
$\epsilon$ is linearly annealed from 1 at the start and to 0 at the end of the training.
No exploration was made in the test steps since $\pi$ is already stochastic. 
A discount factor $\gamma$ is set to $0.9$.

We evaluated several baseline methods and compared the performance with our methods.
\begin{itemize}
    \item \textbf{Random}: This method randomly diffuses drivers to successor roads.
    \item \textbf{Proportional}: This method deploys drivers with the probability proportional to the number of orders in their successor roads. 
    \item \textbf{DRL-based}: This method uses GCN or GAT to approximate $Q$ function in DQN.
\end{itemize}
For DRL-based methods, the followings were compared.
\begin{itemize}
    \item \textbf{$\epsilon$-Greedy} \cite{lin2018efficient}: This method sends all drivers to the next road that has maximum $Q$ value, with the probability $1-\epsilon$, otherwise randomly diffuses drivers. 
    \item \textbf{Entropy} \cite{haarnoja2017reinforcement}: This method is similar to \textbf{Exp}, whereas it uses soft value function ${Q^{\text{soft}}}$ instead of ${Q}$, with the update rule $Q^{\text{soft}}(s,a) = R(s, a, s')+ \gamma / \beta \times \log{\sum_{a' \in \mathcal{A}} \exp  \left({\beta Q^{\text{soft}}(s', a')}\right) }$,
    which employs an entropy term to enforce exploration.
    \item \textbf{Pow / Exp}: This method sends drivers proportional to $Q^\beta$ (power) or $\exp(\beta{Q})$ (exponential) values in \eqref{eq:our_policy_update}.
\end{itemize}

\subsection{Results}
Table \ref{table:real_seoul} shows the order response rate of each proposed method under various conditions.
Total 7 different methods suggested in Section~\ref{sec:setting} are compared.
Note that GAT is used for all of DRL-based methods.
 
To analyze the robustness of our algorithm, we tested with the different number of drivers as in \cite{lin2018efficient}. 
Here, $n\%$ driver ($n=100, 50, 20$) means that we reduced the quantity of drivers including the initial driver distribution and the total driver number at each time step $t$.
We also conducted simulations on different days, since call/driver distribution is highly dependent on the day of the week.
Case (A) shows the result for the same day of the week (with a different date), which has a similar call/driver distribution.
Case (B) shows the result for the different day of the week, which has different call/driver distribution.

\input{table_algos/06_exp_table_1}

In most cases, random dispatch showed the worst result as expected.
The method dispatching proportional to the number of calls was slightly better than the random method.
Overall, RL based methods were shown to outperform non-RL based methods.
A greedy policy update ($\epsilon = 0$, $\beta=\infty$) performed worse than a stochastic policy update, which is consistent with our discussion in Section~\ref{sec:MARL_with_GNN}.
Adding a small portion of randomness (i.e. $\epsilon = 0.1$) as in \cite{lin2018efficient} gave better results but its performance was  inferior to \textbf{Entropy}.
Our proposed method that deploys agents proportional to the function of $Q$ (\textbf{Pow}, \textbf{Exp}) showed the best performance.
We could not find significant difference between \textbf{Pow} and \textbf{Exp}, but \textbf{Pow} gave slightly better result.
Comparing the models used for GNN, GAT performed better than GCN which seems to be because GAT is more expressive than GCN (Table \ref{table:ExpEntropy}).

\input{table_algos/06_exp_table_2}

The comparison of \textbf{Exp} and \textbf{Entropy} given in Table \ref{table:ExpEntropy} provides us interesting insight.
They use the same policy update rule (softmax on $Q$ function), but \textbf{Entropy} method incorporates an entropy term to $Q$ function to encourage exploration.
The experimental result supports our argument that sending too many agents to the same road leads to instant negative feedback on the reward and makes entropy-based exploration superfluous or even detrimental.
Compared to \textbf{Exp}, \textbf{Entropy} showed much dramatic performance degradation as $\beta$ decreases.
We observed that \textbf{Entropy} performed slightly better than \textbf{Exp} when $\beta$ has a large value.
This is because we cannot take advantage of the stochastic policy when $\beta$ is too large, and forced exploration induced by the entropy term becomes somewhat helpful.

\begin{figure}[t!]
\centering
\includegraphics[width=0.48\linewidth]{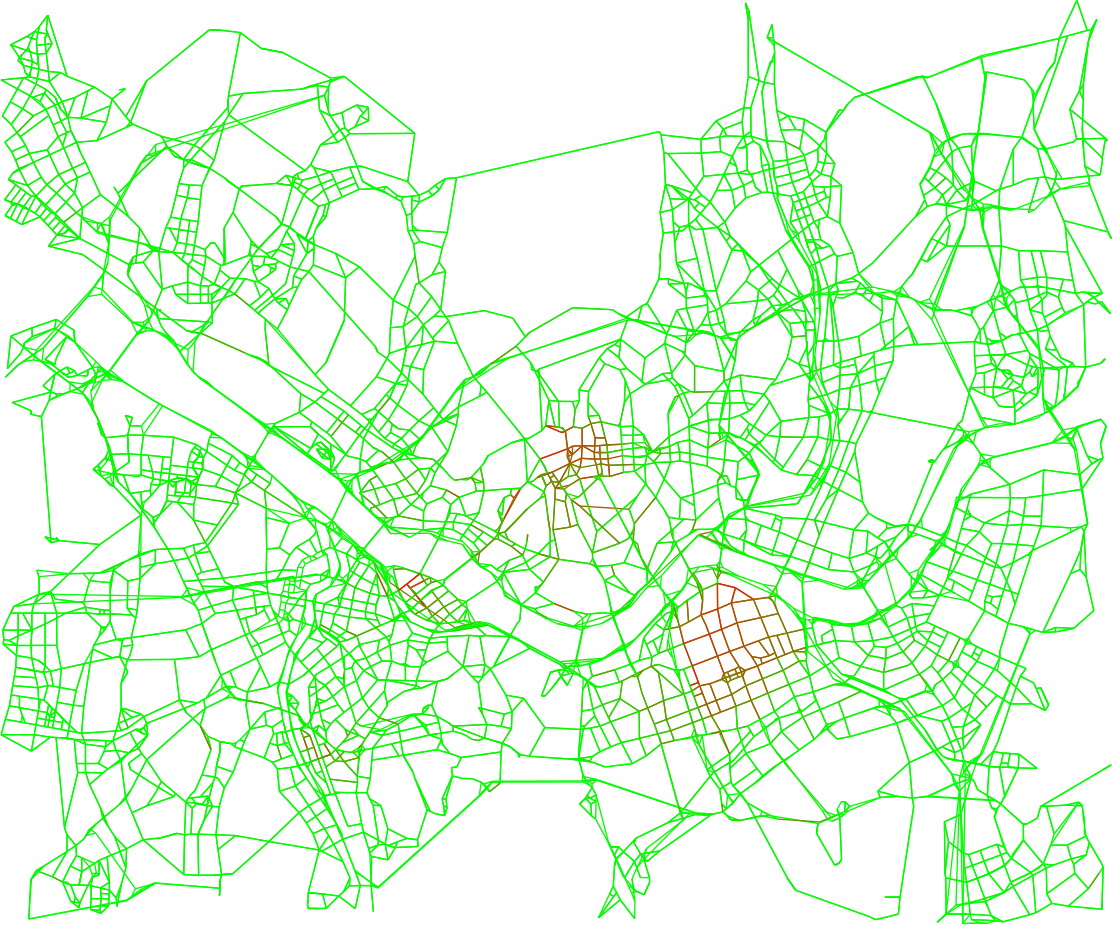}
\includegraphics[width=0.48\linewidth]{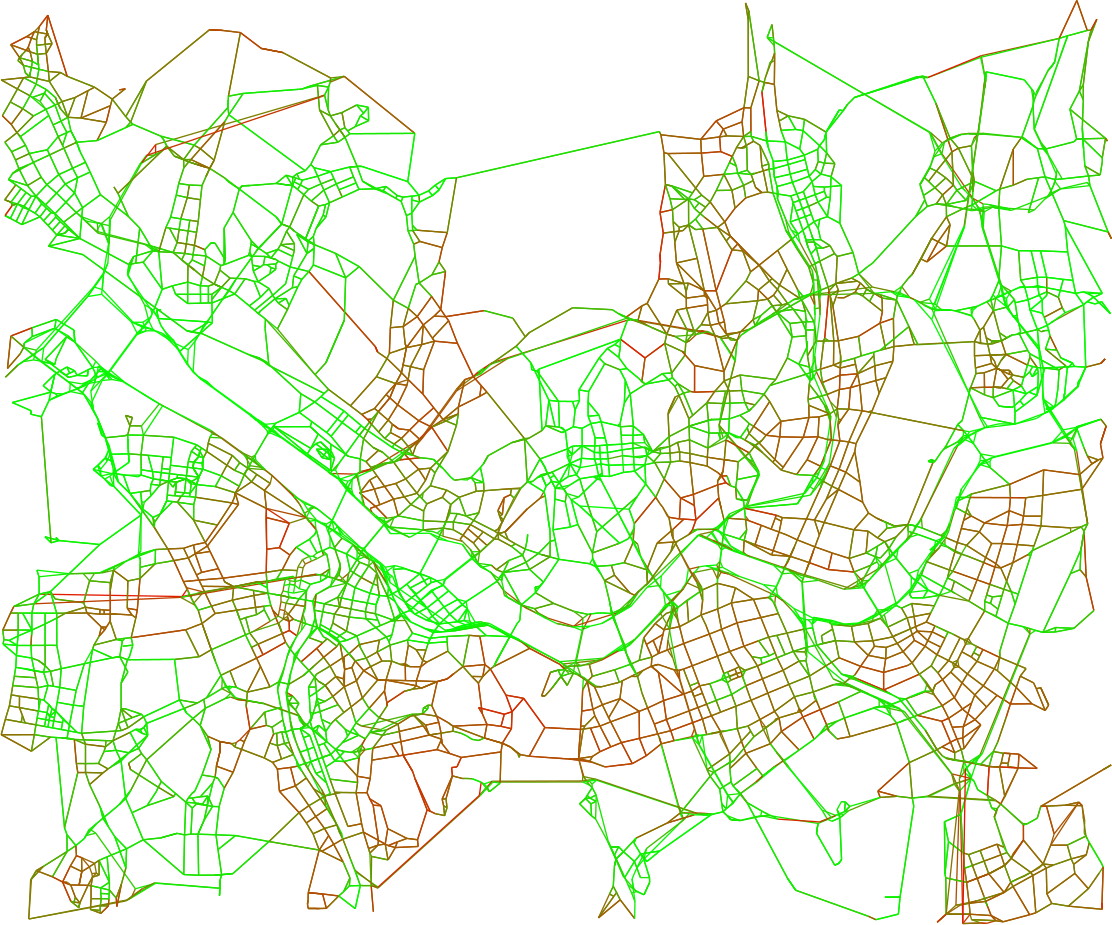}
\caption{Expected $Q$ values of each road in Seoul at 1:00 AM (left) and 8:00 AM (right). A road with more red color has higher $Q$ value.}
\label{fig:qualiativie_result}
\end{figure}

Finally, we report the qualitative result obtained from our work (Fig. \ref{fig:qualiativie_result}).
We plotted the road network that shows the value of $Q$ function for each road.
At 1:00 AM, areas near entertainment districts showed higher $Q$ values.
At 8:00 AM, high values were found throughout the city due to commuting people.
Overall, we could conclude that $Q$ values computed from our model reflect the real situation.

%% file: table_algos/06_exp_table_1.tex
\begin{table}[]
\begin{center}
\caption{Performance comparison under various conditions. $n\%$ means number of drivers. (A) shows result on the same day, different date. (B) shows result on the different day.}
\label{table:real_seoul}
\begin{tabular}{cccc|c|c}
\hline
\multirow{2}{*}{Method} & \multicolumn{5}{c}{Order response rate}\\
 & 100\%  & 50\% & 20\% & 
(A) & (B) \\
\hline\hline
Random       & 0.669 & 0.462 & 0.225 & 0.693 & 0.645 \\
Proportional & 0.694 & 0.490 & 0.238 & 0.708 & 0.665 \\
$\epsilon$-Greedy (${\epsilon = 0.0}$) & 0.685 & 0.505 & 0.269 & 0.646 & 0.626 \\
$\epsilon$-Greedy (${\epsilon = 0.1}$) \cite{lin2018efficient} & 0.752 & 0.541 & 0.280 & 0.741 & 0.718 \\
Entropy (${\beta =20}$) \cite{haarnoja2017reinforcement} & 0.783 & 0.588 & 0.304 & 0.772 & 0.754 \\
Pow (${\beta=3}$) (Ours) & \textbf{0.800} & \textbf{0.601} & \textbf{0.310} & \textbf{0.782} & \textbf{0.771} \\
Exp (${\beta =20}$) (Ours) & 0.791 & 0.592 & 0.305 & 0.779 & 0.765 \\
\hline
\end{tabular}
\end{center}
\end{table}

%% file: table_algos/06_exp_table_2.tex
\begin{table}[]
\caption{Performance comparison of GCN vs. GAT and \textbf{Exp} vs. \textbf{Entropy}. \textbf{Exp} was used for comparing GCN and GAT.
}
\label{table:ExpEntropy}
\begin{center}
\begin{tabular}{ccccccc}
\hline
        & \multicolumn{2}{c}{100\% driver} & \multicolumn{2}{c}{50\% driver} & \multicolumn{2}{c}{20\% driver} \\
$\beta$ & \hspace{0.5em}GCN\hspace{0.5em}             & \hspace{0.58em}GAT\hspace{0.58em}            & \hspace{0.5em}GCN\hspace{0.5em}            & \hspace{0.58em}GAT\hspace{0.58em}            & \hspace{0.5em}GCN\hspace{0.5em}            & \hspace{0.58em}GAT\hspace{0.58em}            \\
\hline
\hline
50      & 0.771           & 0.776          & 0.537          & 0.575          & 0.262          & 0.297          \\
20      & 0.778           & 0.791          & 0.546          & 0.592          & 0.267          & 0.305          \\
10      & 0.775           & 0.782          & 0.545          & 0.583          & 0.267          & 0.300          \\
5       & 0.773           & 0.775          & 0.546          & 0.578          & 0.268          & 0.300    \\
\hline
\end{tabular}
\\
\vspace{1em}
\begin{tabular}{ccccccc}
\hline
     & \multicolumn{2}{c}{100\% driver} & \multicolumn{2}{c}{50\% driver} & \multicolumn{2}{c}{20\% driver} \\
$\beta$ & \hspace{0.7em}Exp\hspace{0.7em}             & Entropy        & \hspace{0.7em}Exp\hspace{0.7em}            & Entropy        & \hspace{0.7em}Exp\hspace{0.7em}            & Entropy        \\
\hline\hline
50   & 0.776         & 0.780          & 0.575         & 0.580         & 0.297         & 0.298         \\
20   & 0.791         & 0.783          & 0.592         & 0.588         & 0.305         & 0.304         \\
10   & 0.782         & 0.759          & 0.583         & 0.563         & 0.300         & 0.293         \\
5    & 0.775         & 0.657          & 0.578         & 0.465         & 0.300         & 0.256    \\
\hline
\end{tabular}
\end{center}
\end{table}

%% file: 07_conclusion.tex
\section{Conclusions}
\label{sec:conclusion}
In this paper, we have presented a novel fleet management strategy in ride-hailing services.
Our approach is distinguished from others by assuming a graph-based spatial condition, which has stronger representative power for road networks than a grid-based condition.
Modified DQN with stochastic policy update is adopted and we showed that GNN can effectively approximate the $Q$ function in our method.
A simulator that reflects real road networks is designed and employed as a training/testing environment to demonstrate the effectiveness of our framework.
Our approach open a new avenue for future research that connects state-of-the-art GNN and MADRL techniques to fleet management problems. 

%% file: 08_appendix.tex
\appendices
\section{Discussion on Stochastic Policy Update}\label{app:example}
In this section, we discuss the convergence and optimality issue of our method.
Due to the complexity of the fleet management problem, it is difficult to conduct a mathematically rigorous convergence analysis.
We can only expect that if an equilibrium state exists for the policy, there will be negative feedback which makes our policy do not deviate from it.
For example, if more drivers are sent to a certain road than the equilibrium state policy, the expected reward will be reduced, and the $Q$ value will also be updated to a smaller value.
Because the policy is proportional to the increasing function of $Q$, we can expect that fewer drivers will be sent next time.

On the optimality issue, we can find a simple counter-example.
Suppose that two roads ($l_1, l_2$) are connected in the same direction and the number of drivers and orders are given by Table. \ref{table:simple_counter_example}.
We assume that all agents are controllable and only two actions are admissible: staying at road 1 (i.e. $l_1 \rightarrow l_1$) and moving to road 2 (i.e. $l_1 \rightarrow l_2$).

\begin{table}[h]
\begin{center}
\caption{Number of drivers and orders.}
\label{table:simple_counter_example}
\begin{tabular}{|c|c|c|}
\hline
                  & Road 1 & Road 2 \\
\hline
Number of drivers ($N_j$) & 10     & 0      \\
Number of orders ($N^{\text{call}}_j$)  & 3      & 7     \\
\hline
\end{tabular}
\end{center}
\end{table}

For simplicity, suppose that we only have a single step and drivers can be distributed in non-integer value.
Then, the reward per unit driver can be expressed as follows:
\begin{equation}
    \mathbb{E}[R_j] = \min \left\{ 1, \frac{N^{\text{call}}_j}{\pi(l_1\rightarrow l_j){N_1}} \right\}, \quad j=1, 2.
\end{equation}
It is clear that an optimal policy is given by $[\pi(l_1 \rightarrow l_1), \pi(l_1 \rightarrow l_2)] = [0.3,0.7]$ and corresponding $Q$ function is given by $[Q(l_1\rightarrow l_1), Q(l_1\rightarrow l_2)] = [1, 1]$.
The optimal total reward for all agents will then be 10.

\begin{figure}[t]
\begin{center}
\includegraphics[width=0.75\linewidth]{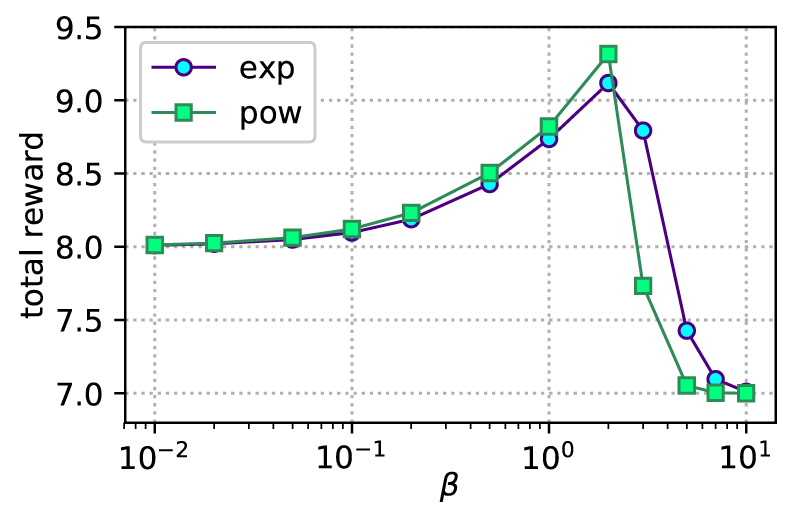}
\end{center}
   \caption{Converged total reward depending on $\beta$ with \textbf{Exp} and \textbf{Pow} $Q$ function.}
\label{fig:pow_exp_result}
\end{figure}

Now consider the stochastic policy update using \eqref{eq:modified_policy_update}.
Assume that initial $Q$ values are set to $[1, 1]$ and $\alpha$ is set to $1$.
Note that if $N_j=0$, we do not update its $Q$ value.
Fig. \ref{fig:pow_exp_result} shows the converged total reward depending on $\beta$.
The resulting policy converges to the uniform distribution $[0.5, 0.5]$ as $\beta \rightarrow 0$ (for both \textbf{Exp} and \textbf{Pow} cases) which results total reward to be 8.
If $\beta \rightarrow \infty$, the stochastic policy converges to the greedy policy $[0, 1]$ and total reward become $7$.
We can observe that the total reward increases from $8$ to nearly $10$ (optimal value) and then decreases to $7$ as $\beta$ increases.
The optimal $\beta$ is approximately $2.0$.
The reason why the reward increases for $\beta \in (0, 2)$ is that our stochastic policy gets more representative power to send more drivers to $l_2$.
The reward decreases for $\beta \in (2, \infty)$ since our policy loses representative power again by becoming excessively greedy.

This example shows that our stochastic policy update does not converge to an optimal policy in general.
But it should be noted here that it still gives better results than the standard greedy policy update rule, which is consistent with our intuition.

\section{Overall Pseudocode}
The pseudocode for the overall learning process is on Algorithm \ref{alg:learning}.

\begin{algorithm}
\caption{Modified DQN with stochastic policy update}\label{alg:learning}
\begin{algorithmic}[1]
\State Initialize $Q$ with $\theta$
\State Initialize $Q'$ with ${\theta}'=\theta$
\For{$m = 1$ to maxiter}
    \State Reset environment and observe initial state $s_0$.
    \For{$t = 0$ to $T$}
        \State Calc $Q^\pi(s_t, l_j; \theta)$ from $s_t$
        \State Calc ${\pi}(l_j \rightarrow l_k|s_t;\theta)$ from $Q^\pi(s_t, l_j; \theta)$
        \State Sample next action $a_t$ from $\pi$
        \State Apply action $a_t$ and observe $R_t, s_{t+1}$
        \State Calc $Q^{\pi'}(s_{t+1}, l_j; {\theta}')$ from $s_{t+1}$
        \State Calc ${\pi}'(l_j \rightarrow l_k |s_{t+1};\theta')$ from $Q^{\pi'}(s_{t+1}, l_j; {\theta}')$
        \For{each idle driver $i=1$ to $N_t$}
            \State Set
            $y_t^i= 
            \begin{dcases}
                R_t^i \; (=1),  & \hspace{-0.5em}\text{if get call}\\
                {\gamma} \hat{Q}^{\pi'}(s_{t+1}, l_{t+1}^i, i;{\theta}'),  &\hspace{-0.5em} \text{otherwise}
            \end{dcases}$
        \EndFor
        \State Update $\theta$ by a gradient descent on loss function
        
        \hspace{1em}$\sum_{i=1}^{N_t}{[ {y_t^i} - Q^\pi(s_t, l^i_{t+1};\theta)]^2}$
        \State Update ${\theta}' \leftarrow \theta$ if needed
    \EndFor
\EndFor
\end{algorithmic}
\end{algorithm}


    
